\documentclass[journal]{IEEEtran}
\usepackage{amsmath,amsfonts}
\usepackage{algorithmic}
\usepackage{algorithm}
\usepackage{array}
\usepackage[caption=false,font=normalsize,labelfont=sf,textfont=sf]{subfig}
\usepackage{textcomp}
\usepackage{stfloats}
\usepackage{url}
\usepackage{verbatim}
\usepackage{graphicx}
\usepackage{multirow}
\usepackage{cite}
\usepackage{booktabs}
\usepackage[capitalize]{cleveref}

\usepackage{colortbl}
\usepackage[table]{xcolor}

\begin{document}

\title{SwinFace: A Multi-task Transformer for \\Face Recognition, Expression Recognition, \\Age Estimation and Attribute Estimation}

\author{Lixiong~Qin, Mei~Wang, Chao~Deng, Ke~Wang, Xi~Chen, Jiani~Hu, Weihong~Deng, ~\IEEEmembership{Member,~IEEE}
	\thanks{Lixiong Qin, Mei Wang, Jiani Hu, Weihong Deng are with the School of Artificial Intelligence, Beijing University of Posts and Telecommunications, Beijing 100876, China (e-mail: lxqin@bupt.edu.cn; wangmei1@bupt.edu.cn; jnhu@bupt.edu.cn; whdeng@bupt.edu.cn).
	Chao Deng, Ke Wang, Xi Chen are with China Mobile Research Institute, Beijing, China (e-mail: dengchao@chinamobile.com; wangkeai@chinamobile.com; chenxiyjy@chinamobile.com).
	}}

\IEEEpubid{\begin{minipage}{\textwidth}\ \\[30pt] \centering
		Copyright \copyright 20xx IEEE. Personal use of this material is permitted. 
		However, permission to use this material for any other purposes must \\ be obtained 
		from the IEEE by sending an email to pubs-permissions@ieee.org.
\end{minipage}}

\maketitle

\begin{abstract}
In recent years, vision transformers have been introduced into face recognition and analysis and have achieved performance breakthroughs. 
However, most previous methods generally train a single model or an ensemble of models to perform the desired task, which ignores the synergy among different tasks and fails to achieve improved prediction accuracy, increased data efficiency, and reduced training time. 
This paper presents a multi-purpose algorithm for simultaneous face recognition, facial expression recognition, age estimation, and face attribute estimation (40 attributes including gender) based on a single Swin Transformer. 
Our design, the SwinFace, consists of a single shared backbone together with a subnet for each set of related tasks. 
To address the conflicts among multiple tasks and meet the different demands of tasks, a Multi-Level Channel Attention (MLCA) module is integrated into each task-specific analysis subnet, which can adaptively select the features from optimal levels and channels to perform the desired tasks. 
Extensive experiments show that the proposed model has a better understanding of the face and achieves excellent performance for all tasks.
Especially, it achieves 90.97\% accuracy on RAF-DB and 0.22 $\epsilon$-error on CLAP2015, which are state-of-the-art results on facial expression recognition and age estimation respectively.
The code and models will be made publicly available at \url{https://github.com/lxq1000/SwinFace}.
\end{abstract}


\section{Introduction}

\IEEEPARstart{F}{ace} recognition and analysis are important topics in the field of computer vision, and have a wide range of applications in security monitoring, digital entertainment, emotion recognition, etc.
Recently, researchers have introduced vision transformers into face recognition and analysis and have achieved performance breakthroughs on some tasks. 
For example, An et al. \cite{Webface260M} proved that ViT\cite{ViT}-based networks can obtain better performance than ResNet\cite{ResNet}-based networks on face recognition; 
TransFER\cite{TransFER} explored transformers for facial expression recognition and achieved significant performance improvement.

However, these transformer-based models are typically designed to achieve only one particular task, which suffers from the following limitations. 
1) Data sparsity. 
Compared with face recognition, face analysis tasks such as facial expression recognition, age estimation, and facial attribute classification are still challenging due to the lack of large available training data, as shown in \cref{tab:datasets}. 
When laser-focused on a single task, big data in face recognition cannot benefit the training of face analysis tasks through knowledge sharing among tasks. 
2) Model efficiency. 
Learning separated networks for different tasks would result in inefficiency in terms of memory and inference speed. 
Since transformers learn general features common to different tasks, they can be shared by face recognition and analysis tasks. 
Although multi-task learning is proposed in convolution neural networks (CNNs) to address these issues, it is still an understudied ﬁeld of research in transformers. 

\begin{figure}[!t]
  \centering
  \includegraphics[width=2.5in]{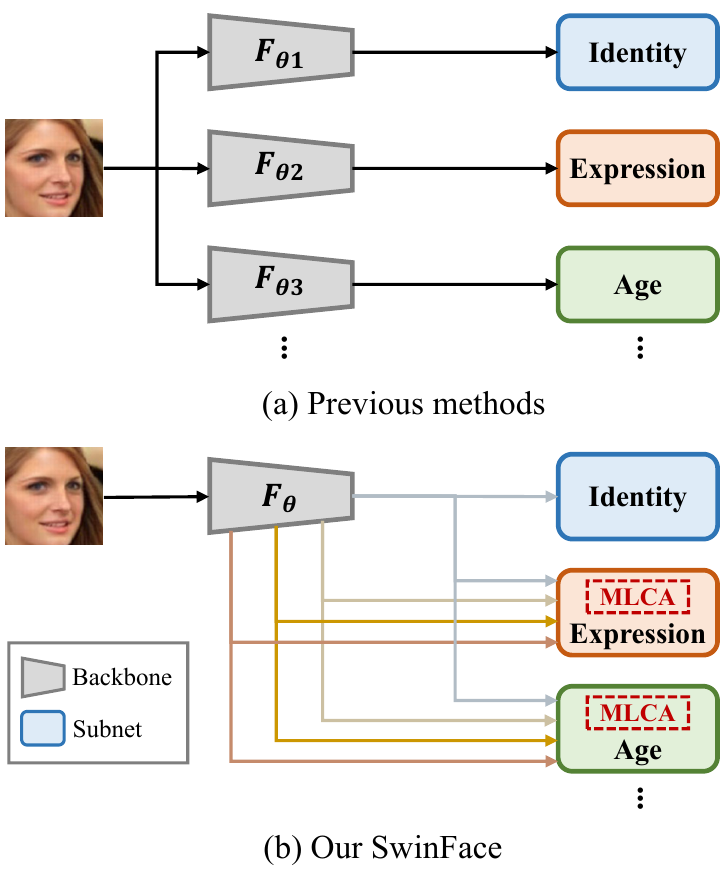}
  \caption{Overview of the previous methods for face recognition and analysis contrasted with our methodology. 
  With shared parameters and the proposed MLCA module, our model can achieve increased application efficiency and improved prediction accuracy.}
  \label{fig:fig1}
\end{figure}

In this paper, we train a transformer jointly in a multi-task learning framework that simultaneously solves the tasks of face recognition, facial expression recognition, age estimation, and face attribute estimation (40 attributes including gender). 
Our design, the SwinFace, consists of a single shared Swin Transformer backbone together with a face recognition subnet and 11 face analysis subnets. 
As shown in \cref{tab:subnets}, to reduce computation, we group some related face analysis tasks and process them by one subnet instead of designing a subnet for each task.
By sharing representations and leveraging the synergy between related tasks, the knowledge contained in a task can be utilized by other tasks, with the hope of improving the performance of all the tasks at hand and reducing the training time.
\cref{fig:fig1} compares the previous methods to our method.

\begin{table}\small
	\caption{Comparison over datasets for face recognition and face analysis.}
	\centering
	\begin{tabular}{ccc}
		\toprule
		Task & Dataset &  Images \\
		\midrule
		Recognition & Webface42M\cite{Webface260M} & 42M\\
		Recognition & MillionCelebs\cite{MillionCelebs} & 18.8M\\
		Recognition & Glint360K\cite{PartialFC-Glint360K} & 17M\\
		Recognition & MS-Celeb-1M\cite{MS-Celeb-1M} & 5.8M\\
		Expression & AffectNet\cite{AffectNet} & 420K\\
		Expression & RAF-DB\cite{DLP-CNN-RAF-DB} & 15,339\\
		Age & IMDB+WIKI\cite{IMDB+WIKI} & 523K\\
		Age & MORPH\cite{MORPH} & 55K\\
		Age & CLAP2015\cite{CLAP2015} & 4699\\
		Attribute & CelebA\cite{CelebA} & 200K\\
		\bottomrule
	\end{tabular}
	\label{tab:datasets}
\end{table}

In addition, multi-task learning is inherently a multi-objective problem and different tasks may conflict. 
For example, the face recognition task learns to extract expression-invariant identity representations whereas the facial expression recognition task is subject-independent, that is, the classification aims to focus on expression but ignores face identity. 
If these two task-specific subnets are simultaneously branched out from the top layer, it is obvious that they will mutually limit performance improvement due to the conflicting targets. 
To avoid this conflict, a Multi-Level Channel Attention (MLCA) module is proposed and integrated into each face analysis subnet, which consists of Multi-Level Feature Fusion (MLFF) and Channel Attention (CA). 
First, MLFF combines features at different levels in an efficient way, which enables our SwinFace to rely on both local and global information of the face to perform face analysis. 
Second, considering that different tasks have different preferences for local and global information, the features at different levels should not be treated equally. 
Therefore, we utilize CA to adaptively assign weights for the features from different levels such that different tasks are allowed to make their own choices for features, which also solves the problem of conflict among tasks to some extent. 
Different from other multi-task learning methods in CNNs, e.g., Hyperface\cite{HyperFace} and AIO\cite{AIO}, which fuse the features from the empirically-selected layers for training the task-specific subnets, our SwinFace adaptively finds the features from appropriate levels, achieving better performance.

This paper makes the following contributions.

\begin{enumerate}
	\item This is the first multi-task model that simultaneously solves a diverse set of face analysis (42 tasks) and recognition tasks using a single transformer.
	\item We propose the Multi-Level Channel Attention (MLCA) module to handle feature extraction conflict of backbone and feature selection of subnets.
	\item The proposed model achieves 90.97\% accuracy on RAF-DB, 0.20 and 0.22 $\epsilon$-error on the validation and test sets of CLAP2015, which are SOTA results on facial expression recognition and age estimation respectively.
\end{enumerate}

\section{Related Work}

\label{sec:formatting}

In this section, related works about face recognition, facial expression recognition, age estimation, face attribute estimation, and multi-task learning are reviewed briefly.

\subsection{Face Recognition}

Face Recognition (FR) is a vital task in computer vision.
In recent years, the development of FR can be summarized in three aspects: loss function, data, and framework.
In terms of loss function, various margin penalties such as SphereFace\cite{SphereFace}, CosFace\cite{CosFace}, and ArcFace\cite{ArcFace} are proposed.
Developing ResNet\cite{ResNet} with IR, ArcFace takes the lead in obtaining a saturation accuracy of 99.83\% on LFW\cite{LFW}.
In terms of data, large-scale training sets \cite{MS-Celeb-1M,PartialFC-Glint360K, MillionCelebs,Webface260M} have been developed.
With the expansion of the dataset, the memory and computing costs linearly scale up to the number of identities in the training set, which calls for a new training framework.
A sparsely updating variant of the FC layer, named Partial FC (PFC)\cite{PartialFC-Glint360K} is invented to save overhead.
To sum up, learning discriminative deep feature embeddings by using million-scale in-the-wild datasets and margin-based softmax loss is the current state-of-the-art approach for FR.

Some researchers have also explored transformer-based FR.
Zhong et al.\cite{FTR} demonstrate that face transformer models trained on MS-Celeb-1M\cite{MS-Celeb-1M} can achieve comparable performance as CNNs with a similar number of parameters.
An et al. \cite{Webface260M} prove that the ViT-based method will be more advantageous on a larger dataset such as WebFace42M.
Existing works rarely explore the facilitation of FR tasks for downstream analysis tasks.
The improvement in descriptive ability brought by large-scale datasets cannot benefit the prediction for expression and age.

\subsection{Facial Expression Recognition}

Facial Expression Recognition (FER) is still a challenging task mainly due to two reasons:
1) Large inter-class similarities. 
Even for the same face identity, a slight change to a small region of the face can determine the expression.
2) Small intra-class similarities.
Samples belonging to the same class may exist large differences in visual appearances, such as skin tone, gender, and age.
Due to such characteristics, although FER generally uses FR to provide initialization for the networks\cite{RAN, SCN, DACL, TransFER}, there is almost no research on the synergy between FER and FR during training.
In existing methods, DLP-CNN\cite{DLP-CNN-RAF-DB} and DACL\cite{DACL} enhance the discriminative ability of expression features by introducing new loss functions.
KTN\cite{KTN} uses adaptive loss re-weight category importance coefficients, alleviating the imbalanced class distribution.
IPA2LT\cite{IPA2LT} and SCN\cite{SCN} address label uncertainties in FER.
gACNN\cite{gACNN}, RAN\cite{RAN} use attention mechanism to adaptively capture the importance of facial regions for occlusion and pose variant FER.
Zhang et al.\cite{ZhangFER} propose an end-to-end deep model for simultaneous facial expression recognition and facial image synthesis, aiming to address the limitation of insufficient training data in improving performance in the field of FER. 
Our method adopts a multi-task learning framework, integrating a more diverse set of tasks to enhance the model's understanding of face, thereby achieving improved performance in facial expression recognition.
AMP-Net\cite{AMPNet} can adaptively capture the diversity and key information from global, local, and salient facial regions. While both approaches employ adaptive feature extraction methods, our motivation lies in mitigating the negative impact of target conflicts between tasks, whereas the motivation behind AMP-Net is to improve the robustness of FER in real-world scenarios.
TransFER\cite{TransFER} first introduces the vision transformer into FER.
This method uses ResNet\cite{ResNet} as stem, reshapes the feature map from stem into a set of tokens, and uses a transformer to model the relationship between these tokens.
Adding a deep ViT to the ResNet stem, the number of parameters of the model reaches 65.2M, which makes it less economical.

\subsection{Age Estimation}

Age Estimation (AE) is an important and very challenging problem in computer vision.
The existing age estimators can be mainly divided into four categories.
1) Classification: DEX\cite{DEX} treats age estimation as a classification problem with 101 classes.
AL\cite{ALRoR} introduces LSTM units to extract local features of age-sensitive regions, improving the age estimation accuracy.
Our model enables the subnet to adaptively select features from different levels, effectively leveraging the local features from the lower level of the backbone.
2) Regression: BridgeNet\cite{BridgeNet} uses local regressors with overlapping subspaces and gating networks with the proposed bridge-tree structure to efficiently mine the continuous relationship between age labels.
3)Ranking: OR-CNN\cite{OR-CNN} and Ranking-CNN\cite{Ranking-CNN} treat ages as ranks.
To estimate a person’s age, they dichotomize whether the person is older than each age or not.
The final estimate is obtained by combining a series of binary classification results.
4) Distribution learning: Age tags are not precise tags and have some ambiguity.
MV\cite{MV} and AVDL\cite{AVDL} robust age estimation using distribution learning.
In addition, DRF\cite{DRF} and DLDLF\cite{DLDLF} recognize that age estimation is a nonlinear regression problem. Both of them connect split nodes to the top layer of convolutional neural networks (CNNs) and deal with inhomogeneous data by jointly learning input-dependent data partitions at the split nodes and age distributions at the leaf nodes.
Current age estimation methods have under-emphasized the importance of face recognition initialization.
These methods\cite{DEX, AGEn, BridgeNet, MV, AVDL, MWR} generally use the large-scale age estimation dataset IMDB-WIKI \cite{IMDB+WIKI} for pre-training, which restricts performance improvement.
Our experiments prove that large-scale face initialization can significantly improve the accuracy of age prediction.
Although our model did not incorporate dedicated structures like DRF\cite{DRF} and DLDLF\cite{DLDLF} to explicitly handle the issue of age inhomogeneity, it still exhibited impressive performance. This outcome further underscores the significance of face recognition initialization and multi-task training frameworks in the context of age estimation.

\subsection{Face Attribute Estimation}

Face attributes give intuitive descriptions of human-comprehensible facial features, such as smile, gender, glasses, beard, etc.
Face Attribute Estimation (FAE) is usually a binary judgment of whether a face has a certain attribute.
Liu et al.\cite{CelebA} released the CelebA dataset, containing about 200K near-frontal images with 40 attributes,
accelerating research in this area \cite{PANDA-1, CelebA,MOON}.
In recent years, some new methods have been continuously proposed to improve the performance of face attribute estimation.
MCNN-AUX\cite{MCNN-AUX} takes advantage of attribute relationships by dividing 40 attributes into nine groups.
MCFA\cite{MCFA} and DMM-CNN\cite{DMM-CNN} exploit the inherent dependencies between face attribute estimation and auxiliary tasks, such as facial landmark localization, improving the performance of face attribute estimation by taking advantage of multi-task learning.
Our method also introduces grouping for the purpose of improving performance and reducing computation.

\subsection{Multi-task Learning}

Multi-task learning (MTL) is first analyzed in detail by Caruana\cite{Multitask}.
In recent years, multi-task learning has been widely applied in computer vision tasks, such as image search \cite{INET}, object detection\cite{MAfasterRCNN}, face recognition\cite{BoostGAN, TSGAN}, and facial analysis\cite{ZhangFER}.
In general, multi-task learning is motivated by the following aspects. 
1) Increased efficiency. 
Multiple tasks can be accomplished simultaneously.
I-Net\cite{INET} jointly performs person re-identification and person search without the need for first detecting and cropping person regions for feature matching.
BoostGAN\cite{BoostGAN} and TSGAN\cite{TSGAN} recover faces from occluded but profile inputs, simultaneously eliminating the impact of pose variation and occlusion, which are two key factors affecting the accuracy of face recognition.
In our method, by combining downstream face analysis tasks into a single model, they can collectively benefit from the advantages of face recognition initialization without the need for multiple sets of backbone network parameters, thereby improving data efficiency and reducing training time.
2) Improved performance.
Zhang et al.\cite{ZhangFER} perform expression synthesis and representation jointly. Both tasks can boost their performance for each other via the unified model.
In our method, a multi-task learning framework can effectively explore synergy among tasks, improving the performance of both face recognition and downstream face analysis tasks.
3) Learning paradigm.
Object detection\cite{MAfasterRCNN} tasks inherently involve multi-task learning, requiring the joint optimization of object classification and bounding box regression. In our work, we do not have the motivation for this aspect.

Our work is primarily inspired by HyperFace and AIO, where the multi-task learning framework can fully explore the relationship between different tasks and enhance the discriminative ability of the shared backbone.
HyperFace\cite{HyperFace} trained an MTL network for face detection, landmarks localization, pose, and gender estimation by fusing the intermediate layers of CNN for improved feature extraction.
AIO\cite{AIO} further expands the function of Hyperface, adds smile detection and age estimation, and demonstrates that analysis tasks benefit from domain-based regularization and network initialization from face recognition task.
AIO has noticed that different tasks have different preferences for features from various levels.
In that method, analysis tasks are divided into two categories: subject-independent and subject-dependent.
AIO believes that subject-independent tasks rely more on local information available from the lower layers of the network, while subject-dependent ones are the opposite.
Under this consideration, the first, third, and fifth convolutional layers are fused for training the subject-independent tasks while subject-dependent tasks are branched out from the sixth convolutional layer.
Unlike HyperFace and AIO, which fuse features from empirically-selected layers for training the task-specific subnets, our method adaptively selects features from the appropriate levels, achieving better performance.

Our ultimate goal is to train a generic model capable of handling all face-related tasks. In addition to the tasks already performed in SwinFace, we also aim to include localization tasks such as pose estimation\cite{zhang2022uncertainty, Pose}, alignment\cite{FAN}, parsing\cite{LaPa}, and 3D reconstruction\cite{chen2021joint, PRNet}. 
For tasks that lack sufficient labeled data, we will consider leveraging semi-supervised learning\cite{tu2023consistent} to enhance the model's performance.


\section{Method}

\begin{figure}[t]
	\centering
	\includegraphics[width=80mm]{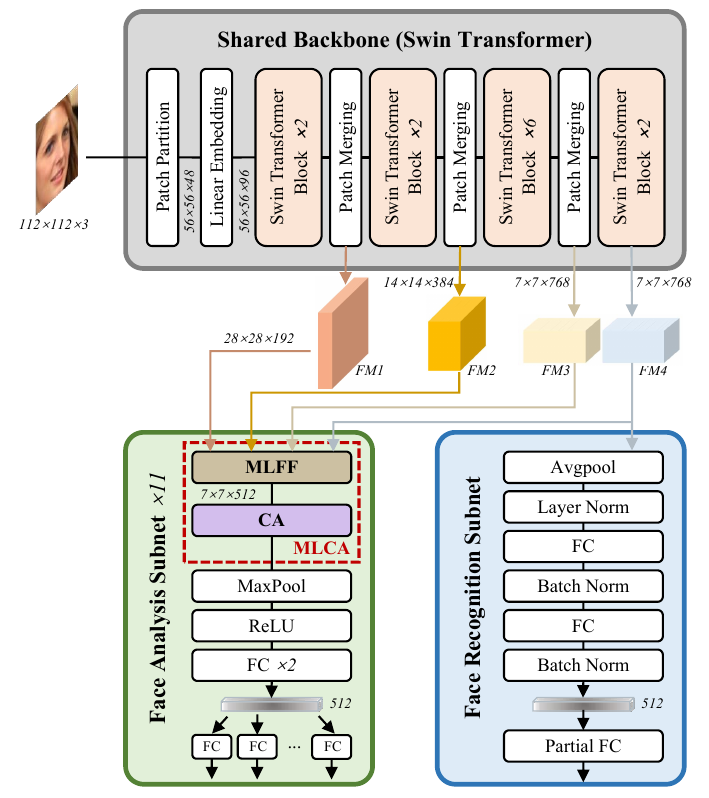}
	
	\caption{Overall architecture for the proposed method.}
	\label{fig:Architecture}
\end{figure}

We proposed a multi-purpose transformer-based model for simultaneous face recognition, facial expression recognition, age estimation, and face attribute estimation.
With a well-designed structure, the model leverages synergy and alleviates target conflict among tasks, resulting in excellent performances.
In this section, we will provide the details of network structure design and training procedure.
\subsection{Overall Architecture}
An overview of the SwinFace architecture is presented in \cref{fig:Architecture}.
In this paper, we adopt a single Swin Transformer\cite{Swin} to extract shared feature maps at different levels.
Based on shared feature maps, we further perform multi-task learning with a face recognition subnet and 11 face analysis subnets.

\subsubsection{Shared Backbone}
The shared Swin Transformer backbone can produce a hierarchical representation. 
The cropped $112\times112\times3$ face image is first split into non-overlapping patches by a patch partition module. 
In our implementation, we use a patch size of $2\times2$ and thus the number of tokens for the subsequent module is $56\times56$ with a dimension of $48$.
A linear embedding layer is applied to this raw-valued feature to project it to 96 dimensions.
After that, the patch merging layers and Swin Transformer blocks are utilized alternately.
The patch merging layers can reduce the number of tokens by a multiple of $2\times2 = 4$, and double the dimension of the tokens.
The Swin Transformer blocks are applied for feature transformation, with the resolution unchanged.
Four feature maps from different levels are finally output for face recognition or analysis tasks.
The scales of these feature maps are $28\times28\times192$, $14\times14\times384$, $7\times7\times768$, and $7\times7\times768$, respectively, and are denoted as FM1 to FM4.

\subsubsection{Face Recognition Subnet}
Face recognition requires robust representations that are not affected by local variations.
Therefore, we only provide the feature map extracted from the top layer, namely FM4, to the face recognition subnet.
Similar to ArcFace\cite{ArcFace}, we introduced the structure that includes BN\cite{BN} to get the final 512-$D$ embedding feature.
Experiments in \cref{ablation} demonstrate that the recognition subnet with FC-BN-FC-BN structure can outperform the counterpart without this structure.

\begin{table}\scriptsize
	\caption{Task assignment for face analysis subnets.
		The output scales of FER, AE, and FAE are 7, 1, and 2, respectively.}
	\centering
	\begin{tabular}{ccc}
		\toprule
		\multicolumn{1}{c}{\multirow{2}{*}{Subnet}}  & \multicolumn{1}{c}{\multirow{2}{*}{Tasks}} & Number \\
		& & of Tasks\\
		\midrule
		Expression & Expression(7), Smiling & 2\\
		\midrule
		Age & Age(1), Young & 2\\
		\midrule
		Gender & Male & 1\\
		\midrule
		\multicolumn{1}{c}{\multirow{2}{*}{Whole}} & Attractive, Blurry, Chubby, Heavy Makeup, & \multicolumn{1}{c}{\multirow{2}{*}{6}} \\
		& Oval Face, Pale Skin &\\
		\midrule
		\multicolumn{1}{c}{\multirow{3}{*}{Hair}} &Bald, Bangs, Black Hair, Blond Hair, Brown Hair,&\multicolumn{1}{c}{\multirow{3}{*}{10}}\\
		&Gray Hair, Receding Hairline, Straight Hair,&\\
		&Wavy Hair, Wearing Hat&\\
		\midrule
		\multicolumn{1}{c}{\multirow{2}{*}{Eyes}} & Arched Eyebrows, Bags Under Eyes, & \multicolumn{1}{c}{\multirow{2}{*}{5}}\\
		&Bushy Eyebrows, Eyeglasses, Narrow Eyes &\\
		\midrule
		Nose & Big Nose, Pointy Nose & 2\\
		\midrule
		\multicolumn{1}{c}{\multirow{2}{*}{Cheek}} & High Cheekbones, Rosy Cheeks, & \multicolumn{1}{c}{\multirow{2}{*}{4}}\\
		&Wearing Earrings, Sideburns &\\
		\midrule
		\multicolumn{1}{c}{\multirow{2}{*}{Mouth}} & 5 o’clock Shadow, Big Lips, Mouth Slightly Open, &\multicolumn{1}{c}{\multirow{2}{*}{6}}\\
		&Mustache, Wearing Lipstick, No Beard &\\
		\midrule
		Chin & Double Chin, Goatee & 2\\
		\midrule
		Neck & Wearing Necklace, Wearing Necktie & 2\\
		\midrule
		Total & & 42\\
		\bottomrule
	\end{tabular}
	\label{tab:subnets}
\end{table}

\subsubsection{Face Analysis Subnets}
The proposed model is able to perform 42 analysis tasks, which are divided into 11 groups according to the relevance of the tasks, as shown in \cref{tab:subnets}.
Tasks in the same group share a face analysis subnet to reduce computation.
Each subnet consists of a Multi-Level Channel Attention (MLCA) module, a max pooling layer, a ReLU activation layer, two consecutive fully connected layers, and a series of fully connected layers for output.
MLCA is the critical structure that enables subnets to make their own choices for features solving the problem of conflict among tasks to some extent.

\subsection{Multi-Level Channel Attention}
Conventional face recognition and analysis methods branch out task-specific subnets only from the top layer of the backbone.
Tasks with conflicting targets will therefore mutually limit performance improvement.
To solve this issue, we propose Multi-Level Channel Attention (MLCA) module and integrate it into each face analysis subnet.
The MLCA module consists of a Multi-Level Feature Fusion (MLFF) module and a Channel Attention (CA) module.
MLFF is used to combine feature maps at different levels enabling the task-specific subnet to rely on both local and global information of the faces and CA emphasizes the contributions of different levels for the specific group of tasks.
It is important to note that previous methods primarily utilized feature fusion and adaptation to enhance the robustness of features in single-task scenarios. 
For instance, in the AMP-Net\cite{AMPNet} for FER, the Gate-OSA, a feature fusion module, is employed in the GP module to learn facial features from diverse receptive fields. The LP and AP modules leverage CBAM\cite{CBAM}, a feature adaptation module, to further enhance the extracted features.
In contrast, the primary motivation behind the proposed MLCA is to alleviate target conflicts among tasks in the multi-task scenario.

\subsubsection{Multi-Level Feature Fusion}
\begin{figure}[t]
	\centering
	\includegraphics[width=80mm]{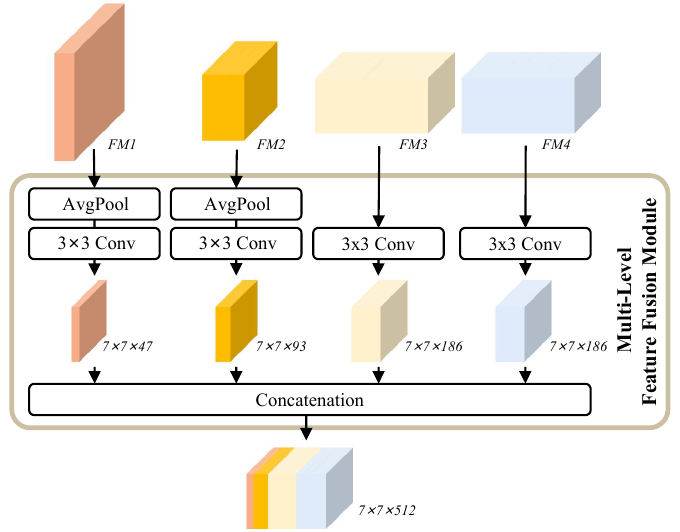}
	
	\caption{Multi-Level Feature Fusion module.}
	\label{fig:MLFF}
\end{figure}

Swin Transformer has a hierarchical architecture, which is a stack of multiple transformer blocks.
Blocks at lower levels capture low-level elements such as basic colors and edges, while the ones at higher levels encode abstract and semantic cues.
It is natural to combine the features from different levels for analysis tasks.
In doing so,  the individual analysis task is enabled to rely on both local and global information.
As shown in \cref{fig:MLFF}, to keep the scale of the feature maps from each level consistent, the FM1 and FM2 are first down-sampled by average pooling.
Then, 4 independent $3\times3$ convolutions are applied to the input feature maps of FM1-4 to proportionally reduce the number of channels.
We further concatenate them in the channel dimension to get a 512-dimensional feature map.
The MLFF module enables the transformer blocks in the backbone to be directed both by the successive blocks as well as the neighboring face analysis subnets during training, which can speed up the convergence of the model and also improves the generalization of the extracted features.

\subsubsection{Channel Attention}
\begin{figure}[t]
	\centering
	\includegraphics[width=80mm]{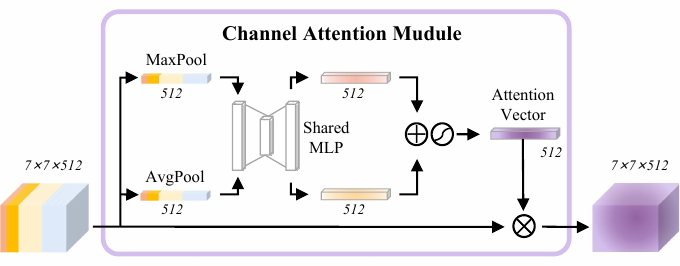}
	
	\caption{Channel Attention module.}
	\label{fig:CA}
\end{figure}

Different analysis tasks have different preferences for local and global information.
MLFF combines features at different levels in an efficient way and provides a 512-dimensional concatenated feature map.
We hope to separately emphasize the contributions of different channels in the feature map with an attention vector, in which the $i$-th activation value of the attention vector corresponds to the $i$-th channel of the feature map.
For this motivation, in our Channel Attention (CA) module, we follow CBAM\cite{CBAM} to calculate the attention vector, as shown in \cref{fig:CA}.
First, the average-pooled and max-pooled features are obtained from the concatenated feature map as two different spatial context descriptors.
Both descriptors are then forwarded to a shared multi-layer perceptron (MLP) with one hidden layer.
The output feature vectors of the two descriptors are merged using element-wise summation, and then the attention vector is obtained through a sigmoid function.
The input concatenated feature map and the attention vector are element-wise multiplied to obtain the channel-weighted feature map as the output of the CA module.

\subsection{Training}
We use a multi-task learning framework so that the model can simultaneously solve the tasks of face recognition, facial expression recognition, age estimation, and face attribute estimation.
As shown in \cref{tab:analysis-datasets}, the training sets can be divided into four categories according to the provided labels.
We start with a pre-trained model which only includes a Swin Transformer backbone and a face recognition subnet.
The Swin Transformer backbone is then shared by the face recognition task and all 42 face analysis tasks. 
The loss functions and training datasets are illustrated as follows.

\begin{table}\small
	\caption{Training datasets for multi-task training phase can be divided into four categories by label types.}
	\centering
	\begin{tabular}{cc}
		\toprule
		Label & Dataset  \\
		\midrule
		Identity & MS-Celeb-1M\cite{MS-Celeb-1M}\\
		Expression & RAF-DB\cite{DLP-CNN-RAF-DB} AffectNet\cite{AffectNet}\\
		Age Gender & IMDB+WIKI\cite{IMDB+WIKI} Adience\cite{Adience} MORPH\cite{MORPH}\\
		Attribute & CelebA\cite{CelebA}\\
		\bottomrule
	\end{tabular}
	\label{tab:analysis-datasets}
\end{table}

\subsubsection{Face Recognition}
We train the task of face recognition on the large-scale face recognition dataset MS-Celeb-1M\cite{MS-Celeb-1M} with CosFace\cite{CosFace}:
	\begin{equation}
		L_R=-\frac{1}{N_R}\sum_{i=1}^{N_R}\log\frac{e^{s(\cos\theta_{y_i}-m)}}  {e^{s(\cos \theta_{y_i}-m)}+ {\textstyle \sum_{j=1,j\neq y_i}^{n}e^{s \cos \theta _j}}}.
		\label{eq:recognition-loss}
	\end{equation}
Assume that the weight of the last fully connected layer is written as $W \in \mathbb{R}^{d\times  n}$, where $n$ is the number of identities.
We use $W_j\in \mathbb{R}^d$ to denote the $j$-th column of the weight $W$
and $x_i \in \mathbb{R}^d$ to denote the deep feature of the $i$-th sample, belonging to the $y_i$-th class.
$\theta_j$ is the angle between the weight $W_j$ and the feature $x_i$.
The embedding feature $\left \| x_i \right \|$  is fixed by $l_2$ normalization and re-scaled to $s$.
$m$ is the CosFace margin penalty.
In our implementation, s is set to 64, and m is set to 0.4.
$N_R$ is the number of samples with identity labels in each training batch.
In addition, we introduce PFC\cite{PartialFC-Glint360K} to conserve computing resources.
Sampling ratio $r$ is set to 0.3.
Experiments in \cref{ablation} demonstrate that CosFace\cite{CosFace} is more suitable for our model than other loss functions such as ArcFace\cite{ArcFace}.

\subsubsection{Facial Expression Recognition} 
It is a multi-classification problem.
Due to the lack of a large-scale training set, we merge AffectNet\cite{AffectNet} and RAF-DB\cite{DLP-CNN-RAF-DB} by labels for basic expression analysis.
The expressions include surprise, fear, disgust, happiness, sadness, anger, and neutral.
The loss function for training is as follows:
\begin{equation}
	L_E=-\frac{1}{N_E}\sum_{i=1}^{N_E}\left [ \sum_{c=1}^{7} y_{ic}\log(p_{ic})   \right ],
	\label{eq:expression-loss}
\end{equation}
where $y_{ic} = 1$ if the $i$-th sample belongs to expression class $c$, otherwise 0.
The predicted probability that the $i$-th sample belongs to expression class $c$ is given by $p_{ic}$.
$N_E$ is the number of samples with expression labels in each training batch.

\subsubsection{Age Estimation}
We formulate the age estimation task as a regression problem.
IMDB+WIKI\cite{IMDB+WIKI}, Adience\cite{Adience} and MORPH\cite{MORPH} are used for training. 
we use a linear combination of Gaussian loss and Euclidean loss following Ranjan et. al.\cite{AIO}:

\begin{footnotesize}
	\begin{equation}
		L_A=\frac{1}{N_A}\sum_{i=1}^{N_A}\left [  (1-\lambda)\frac{1}{2}(\hat{a_i}-a_i )^2+\lambda \left ( 1-\exp (-\frac{(\hat{a_i}-a_i )^2}{2\sigma ^2} ) \right )    \right ],
		\label{eq:age-loss}
	\end{equation}
\end{footnotesize}
where $\hat{a_i}$ is the predicted age for sample $x_i$, $a_i$ is the ground-truth age and $\sigma$ is the standard deviation of the annotated age value.
$\lambda$ is initialized with 0 at the start of the training, and increased to 1 subsequently.
$\sigma$ is fixed at 3 if not provided by the training set.
$N_A$ is the number of samples with age labels in each training batch.

\subsubsection{Face Attribute Estimation} 
Face attribute estimation consists of 40 binary classification problems and is trained using CelebA\cite{CelebA}.
Especially, the training of gender recognition also uses labels from IMDB+WIKI\cite{IMDB+WIKI}, Adience\cite{Adience} and MORPH\cite{MORPH}.
The loss function for a single FAE task is as follows:
\begin{equation}
	L_{A_j}=\frac{1}{N_{A_j}}\sum_{i=1}^{N_{A_j}}\left [ -(1-q_i)\cdot \log(1-p_i)-q_i\cdot\log (p_i)   \right ],
	\label{eq:smile-loss}
\end{equation}
where $q_i = 1$ for the $j$-th attribute exists and 0 otherwise.
$p_i$ is the predicted probability that the $i$-th input face contains the $j$-th attribute.
$N_{A_j}$ is the number of samples with the $j$-th attribute labels in each training batch.

\subsubsection{Total Loss function}
The final overall loss $L$ is the weighted sum of individual loss functions, given in \eqref{eq:total-loss}:
\begin{equation}
	L_{total} = \sum_{t\in T}^{}\alpha _tL_t,
	\label{eq:total-loss}
\end{equation}
where $T=\{R, E, A, A_1, A_2, ..., A_{40}\}$ represents tasks and $\alpha_t$ is the weight of task $t$.
The loss-weights are respectively set to 1.0 in experience.

\section{Experiments}

\subsection{Datasets}

\subsubsection{Face Recognition}
We use MS-Celeb-1M\cite{MS-Celeb-1M} for pre-training and multi-task training. 
MS-Celeb-1M is one of the most popular large-scale training databases for face recognition and we use the clean version refined by ArcFace\cite{ArcFace}, which contains 5.8M images of 85,742 celebrities.
For testing, we report the verification performance of models on several mainstream benchmarks including LFW\cite{LFW}, CFP-FP\cite{CFP-FP}, AgeDB-30\cite{AgeDB-30}, CALFW\cite{CALFW} , CPLFW\cite{CPLFW}, and IJB-C\cite{IJB-C} databases.
LFW database contains 13,233 face images from 5,749 different identities, which is a classic benchmark for unconstrained face verification.
CFP-FP and CPLFW are built to emphasize the cross-pose challenge while AgeDB-30 and CALFW are built for the cross-age challenge.
IJB-C contains faces with extreme viewpoints, resolution, and illumination, which makes it more challenging.

\subsubsection{Facial Expression Recognition}
RAF-DB \cite{DLP-CNN-RAF-DB} is a real-world expression dataset containing 12,271 training and 3,068 test images for basic expression analysis. 
AffectNet \cite{AffectNet} is the largest publicly available FER dataset so far, containing 420K images with manually annotated labels.
RAF-DB provides more accurate labels, but with a smaller sample amount, while AffectNet is just the opposite.
We merge AffectNet and RAF-DB by labels for boosting performance.
We report the overall accuracy on the RAF-DB test set.

\subsubsection{Age Estimation}
We use IMDB+WIKI\cite{IMDB+WIKI}, Adience\cite{Adience}, and MORPH\cite{MORPH} for training.
IMDB+WIKI, which contains 523K images in total, is the largest dataset for age estimation, where the images are crawled from celebrities on IMDb and Wikipedia websites.
Adience contains 26,580 images across 2,284 subjects with a label from eight different age groups.
We take the average age of each group as the regression label.
MORPH is the largest database with precise age labeling and ethnicities, including about 55K face images and age ranges from 16 to 77 years.
The Chalearn LAP challenge\cite{CLAP2015} is the first competition for apparent age estimation, collecting 2476 images for training, 1136 images for validation, and 1079 images for testing. 
The dataset offers the standard deviation for each age label.
After finetuning the age subnet, we report age estimation performance on both the validation and test split.

\subsubsection{Face Attribute Estimation}
CelebA\cite{CelebA} consists of 162,770 images for training, 19,867 images for validation, and 19,962 images for testing. 
We report the FAE performance on the testing split.
In particular, the training of gender recognition also uses labels from IMDB+WIKI\cite{IMDB+WIKI}, Adience\cite{Adience}, and MORPH\cite{MORPH}.

For data prepossessing, we follow the recent papers \cite{SphereFace,CosFace,ArcFace} to generate the aligned face crops $(112 \times 112)$.
We perform face alignment using affine transform and matrix rotation in OpenCV.
For facial images without key-point labels, MTCNN\cite{MTCNN} is used to collect landmarks.

\begin{table*}[!ht]\small
	\caption{Comparison for face recognition models.
		Number of backbone parameters of face recognition models. The 1:1 verification accuracy on the LFW\cite{LFW}, CFP-FP\cite{CFP-FP}, AgeDB-30\cite{AgeDB-30}, CALFW\cite{CALFW} and CPLFW\cite{CPLFW} and IJB-C\cite{IJB-C} datasets. }
	\centering
	\begin{tabular}{|c|c|c|c|c|c|c|c|c|c|c|c|c|}\hline
		\multicolumn{1}{|c|}{\multirow{2}{*}{Method}}& \multicolumn{1}{|c|}{Params} & \multicolumn{5}{c|}{Verification Accuray} & \multicolumn{6}{c|}{IJB-C TAR@FAR}\\ \cline{3-13}
		& (M) & { \scriptsize LFW} & { \scriptsize CFP-FP} & { \scriptsize AgeDB-30} & { \scriptsize CALFW} & { \scriptsize CPLFW} & 1e-6 & 1e-5 & 1e-4 & 1e-3 & 1e-2 & 1e-1 \\ \hline
		
		ResNet-50\cite{ResNet} & 43.6 & 99.69 & \underline{98.14} & 97.53 & 95.87 & 92.45 & 81.43 & 90.98 & 94.32 & 96.38 & 97.82 & 98.75\\ \hline
		ViT\cite{ViT} & 63.2 & \underline{99.83} & 96.19 & 97.82 & 95.92 & 92.55 & - & - & 95.96 & 97.28 & 98.22 & 98.99 \\ 
		V2T-ViT\cite{T2T-ViT} & 63.5 & 99.82 & 96.59 & \underline{98.07} & 95.85 & 93.00 & - & - & 95.67 & 97.10 & 98.14 & 98.90 \\ 
		ViT-P10S8\cite{FTR} & 63.5 & 99.77 & 96.43 & 97.83 & 95.95 & 92.93 & - & - & 96.06 & 97.45 & 98.23 & 98.96 \\ 
		ViT-P12S12\cite{FTR} & 63.5 & 99.80 & 96.77 & 98.05 & \textbf{96.18} & \underline{93.08} & - & - & \underline{96.31} & \underline{97.49} & \underline{98.38} & \underline{99.04} \\ \hline
		
		Swin-T\cite{Swin} & 28.5 & 99.80 & 97.91 & 97.85 & 95.98 & 92.60 & \underline{88.54} & \underline{93.71} & 95.75 & 97.13 & 98.01 & 98.86 \\ 
		\textbf{SwinFace} & 28.5 & \textbf{99.87} & \textbf{98.60} & \textbf{98.15} & \underline{96.10} & \textbf{93.42} & \textbf{90.82} & \textbf{94.93} & \textbf{96.73} & \textbf{97.79} & \textbf{98.43} & \textbf{99.08} \\ \hline
		
	\end{tabular}
	\label{tab:face-recognition-comparison}
\end{table*}

\subsection{Implementation Details}
The Swin Transformer backbone adopts the tiny version (Swin-T) which includes 28.5M parameters.
The face recognition subnet has about 1M parameters excluding PFC and each face analysis subnet has about 3.5M parameters.
The model is trained on 4 NVIDIA Tesla T4 GPUs.

We first pre-train the Swin Transformer backbone and face recognition subnet for robust face recognition initialization.
For multi-task learning, we load the pre-trained backbone and face recognition subnet and randomly initialize 11 face analysis subnets.

\subsubsection{Pre-training}
We employ an AdamW\cite{AdamW} optimizer for 40 epochs using a cosine decay learning rate scheduler and 5 epochs of linear warm-up. 
A batch size of 512, an initial learning rate of $5\times10^{-4}$, a warm-up learning rate of $5\times10^{-7}$, a minimum learning rate of $5\times10^{-6}$, and a weight decay of 0.05 are used. 
Data augmentation includes horizontal flip augmentation only.

\subsubsection{Multi-task Learning}
The training lasts 80k steps, of which 8k steps are warm-up steps. 
We set the number of samples from each of the four categories (shown in \cref{tab:analysis-datasets}) to be 128 at each step. 
(Note that there are thus 256 samples per step for training the task of gender recognition.)
For face recognition samples, Only a horizontal flip is utilized.
For other samples, data augmentation includes horizontal flip, Randaugment\cite{Randaugment}, and Random Erasing\cite{RandomErasing}
to alleviate the lack of variations.
Other settings are the same as the pre-training phase.

\subsection{Performance Evaluation}

\subsubsection{Face Recognition}
\cref{tab:face-recognition-comparison} gives the comparison between SwinFace and other face recognition models based on ResNet\cite{ResNet} and Transformers\cite{ViT, T2T-ViT, FTR}.
The performance of models based on transformers is quoted from \cite{FTR}.
These models are trained on MS-Celeb-1M\cite{MS-Celeb-1M}, so a fair comparison can be achieved.
The results prove that SwinFace outperforms other models in almost all test protocols, although the number of parameters of its backbone is much smaller than other models.
In particular, The SwinFace outperforms the Swin-T model on all benchmarks for face recognition which shows that multi-task learning can enhance face recognition capabilities.

\begin{table}\small
	\caption{Comparison for facial expression recognition on RAF-DB\cite{DLP-CNN-RAF-DB}.}
	\centering
	\begin{tabular}{cc}
		\toprule
		Method & Accuray \\
		\midrule
		DLP-CNN\cite{DLP-CNN-RAF-DB} & 80.89\\
		gACNN\cite{gACNN} & 85.07\\
		IPA2LT\cite{IPA2LT} & 86.77\\
		RAN\cite{RAN} & 86.90\\
		CovPool\cite{CovPool} & 87.00\\
		SCN\cite{SCN} & 87.03\\
		DACL\cite{DACL} & 87.78\\
		KTN\cite{KTN} & 88.07\\
		Zhang et al.\cite{ZhangFER} & 89.01\\
		AMP-Net\cite{AMPNet} & 89.25 \\
		TransFER\cite{TransFER} & \underline{90.91}\\
		\textbf{SwinFace} & \textbf{90.97}\\
		\bottomrule
	\end{tabular}
	\label{tab:expression}
\end{table}

\subsubsection{Facial Expression Recognition}
As shown in \cref{tab:expression}, we compare SwinFace with the state-of-the-art methods on RAF-DB\cite{DLP-CNN-RAF-DB}.
Our method outperforms state-of-the-art methods, resulting in an accuracy of 90.97\%.
It is noting that some methods such as SCN [27] and KTN [17] achieve the reported performance by applying trivial loss functions, while our method achieves better performance with the standard cross-entropy loss only.
Compared to AMP-Net\cite{AMPNet}, our approach does not require complex feature enhancement structures for each region. We achieve superior results solely through the use of a simple MLCA module.
Furthermore, to obtain the reported performance, TransFER\cite{TransFER} experimentally determines that features should be extracted from the third stage of the backbone and includes 65.2M parameters.
Allowing the expression subnet to adaptively extract features from appropriate levels, our method achieves higher accuracy with a total of 32M parameters in the Swin Transformer backbone and the expression subnet, 
which proves the great benefits of multi-task learning in improving accuracy and efficiency.

\begin{table}\small
	\caption{Comparison for age estimation on CLAP2015\cite{CLAP2015}.}
	\centering
	\begin{tabular}{c|cc|cc}
		\toprule
		
		\multicolumn{1}{c|}{\multirow{2}{*}{Method}}&  \multicolumn{2}{c|}{Validatation} & \multicolumn{2}{c}{Test}\\ 

		& MAE & $\epsilon$-error & MAE &  $\epsilon$-error  \\
		\midrule
		AIO\cite{AIO} & - & 0.29 & - & -\\
		AgeNet\cite{AgeNet} & 3.33 & 0.29 & - & 0.26\\
		DEX\cite{DEX} & 3.25 & 0.28 & - & 0.26\\
		AGEn\cite{AGEn} & 3.21 & 0.28 & 2.94 & 0.26\\
		AL-RoR\cite{ALRoR} & 3.14 & 0.27 & - & \underline{0.25} \\
		BridgeNet\cite{BridgeNet} & 2.98 & \underline{0.26} & 2.87 & 0.26\\
		MWR\cite{MWR} & \underline{2.95} & \underline{0.26} & \underline{2.77} & \underline{0.25}\\
		\textbf{SwinFace} & \textbf{2.50} & \textbf{0.20} & \textbf{2.47} & \textbf{0.22}\\
		\bottomrule
	\end{tabular}
	\label{tab:age}
\end{table}

\subsubsection{Age Estimation}
As shown in \cref{tab:age}, we evaluate age estimation task on CLAP2015 using the metrics of MAE and $\epsilon$-error.
For each face, CLAP2015 provides the standard deviation of age values by multiple annotators.
$\epsilon$-error is defined as $1-\exp(-\frac{(\hat{a}-a)^2}{2\sigma^2})$,
where $\sigma$ is the standard deviation of the sample, $\hat{a}$ is the predicted age, and $a$ is the ground-truth age.
The average $\epsilon$-error over all test images is reported.
Both validation and test splits of CLAP2015 are used.
For evaluation on the validation set, we use the training set to finetune the age subnet.
For evaluation on the test set, we use the validation set, as well as the training set, to
finetune the age subnet, as in \cite{AgeNet, AGEn, BridgeNet, MWR}.
For both splits, the finetuning lasts 4k steps, without warm-up steps. A minimum learning rate of $5\times10^{-7}$ is used. Data augmentation includes horizontal flip, Randaugment\cite{Randaugment}, and Random Erasing\cite{RandomErasing}to alleviate the lack of variations. Other settings are the same as the Swin Transformer's pre-training phase.
Conventional algorithms\cite{BridgeNet, MWR} introduce complex mechanisms to improve performance.
However, only applying a simple subnet with MLCA, SwinFace outperforms all conventional algorithms.
Significant MAE margin of 0.45 (0.30) and $\epsilon$-error margin of 0.06 (0.03) are achieved on the validation (test) split.

\begin{table*}[!ht] \footnotesize
	\caption{Comparison for face attribute estimation on CelebA\cite{CelebA}.}
	\centering
	\begin{tabular}{|c|c|c|c|c|c|c|c|c|c|c|c|c|c|c|c|c|c|c|c|c|c|}\hline
		& \rotatebox{90}{5 o’clock Shadow} & \rotatebox{90}{Arched Eyebrows} & \rotatebox{90}{Attractive} & \rotatebox{90}{Bags Under Eyes} & \rotatebox{90}{Bald} & \rotatebox{90}{Bangs} & \rotatebox{90}{Big Lips} & \rotatebox{90}{Big Nose} & \rotatebox{90}{Black Hair} & \rotatebox{90}{Blond Hair}
		& \rotatebox{90}{Blurry} & \rotatebox{90}{Brown Hair} & \rotatebox{90}{Bushy Eyebrows} & \rotatebox{90}{Chubby} \\
		\midrule
		PANDA-1\cite{PANDA-1} & 88.00 & 78.00 & 81.00 & 79.00 & 96.00 & 92.00 & 67.00 & 75.00 & 85.00 & 93.00 & 86.00 & 77.00 & 86.00 & 86.00\\
		LNets+ANet\cite{CelebA} & 91.00 & 79.00 & 81.00 & 79.00 & 98.00 & 95.00 & 68.00 & 78.00 & 88.00 & 95.00 & 84.00 & 80.00 & 90.00 & 91.00\\
		MOON\cite{MOON} & 94.03 & 82.26 & 81.67 & 84.92 & 98.77 & 95.80 & 71.48 & 84.00 & 89.40 & 95.86 & 95.67 & 89.38 & 92.62 & 95.44\\
		NSA\cite{NSA} & 93.13 & 82.56 & 82.76 & 84.86 & 98.03 & 95.71 & 69.28 & 83.81 & 89.03 & 95.76 & 95.96 & 88.25 & 92.66 & 94.94\\
		MCNN-AUX\cite{MCNN-AUX} & 94.51 & 83.42 & 83.06 & 84.92 & 98.90 & 96.05 & 71.47 & 84.53 & 89.78 & 96.01 & 96.17 & 89.15 & 92.84 & 95.67\\
		MCFA\cite{MCFA} & 94.00 & 83.00 & 83.00 & 85.00 & 99.00 & 96.00 & 72.00 & 84.00 & 89.00 & 96.00 & 96.00 & 88.00 & 92.00 & 96.00\\
		DMM-CNN\cite{DMM-CNN} & 94.84 & 84.57 & 83.37 & 85.81 & 99.03 & 96.22 & 72.93 & 84.78 & 90.50 & 96.13 & 96.40 & 89.46 & 93.01 & 95.86\\
		\textbf{SwinFace} & 94.60 & 83.91& 82.61 & 84.24 & 98.99 & 96.09 & 71.26 & 83.98 & 90.17 & 95.94 & 96.04 & 89.11 & 92.62 & 95.69 \\
		\midrule
		& \rotatebox{90}{Double Chin} & \rotatebox{90}{Eyeglasses} & \rotatebox{90}{Goatee} & \rotatebox{90}{GrayHair} & \rotatebox{90}{Heavy Makeup} & \rotatebox{90}{High Cheekbones} & \rotatebox{90}{Male} & \rotatebox{90}{Mouth Slightly Open} & \rotatebox{90}{Mustache} & \rotatebox{90}{Narrow Eyes} & \rotatebox{90}{No Beard} & \rotatebox{90}{Oval Face} & \rotatebox{90}{Pale Skin} & \rotatebox{90}{Pointy Nose} \\
		
		\midrule
		PANDA-1\cite{PANDA-1} & 88.00 & 98.00 & 93.00 & 94.00 & 90.00 & 86.00 & 97.00 & 93.00 & 93.00 & 84.00 & 93.00 & 65.00 & 91.00 & 71.00\\
		LNets+ANet\cite{CelebA} & 92.00 & 99.00 & 95.00 & 97.00 & 90.00 & 88.00 & 98.00 & 92.00 & 95.00 & 81.00 & 95.00 & 66.00 & 91.00 & 72.00\\
		MOON\cite{MOON} & 96.32 & 99.47 & 97.04 & 98.10 & 90.99 & 87.01 & 98.10 & 93.54 & 96.82 & 86.52 & 95.58 & 75.73 & 97.00 & 76.46\\
		NSA\cite{NSA} & 95.80 & 99.51 & 96.68 & 97.45 & 91.59 & 87.61 & 97.95 & 93.78 & 95.86 & 86.88 & 96.17 & 74.93 & 97.00 & 76.47\\
		MCNN-AUX\cite{MCNN-AUX} & 96.32 & 99.63 & 97.24 & 98.20 & 91.55 & 87.58 & 98.17 & 93.74 & 96.88 & 87.23 & 96.05 & 75.84 & 97.05 & 77.47\\
		MCFA\cite{MCFA} & 96.00 & 100.00 & 97.00 & 98.00 & 92.00 & 87.00 & 98.00 & 93.00 & 97.00 & 87.00 & 96.00 & 75.00 & 97.00 & 77.00\\
		DMM-CNN\cite{DMM-CNN} & 96.39 & 99.69 & 97.63 & 98.27 & 91.85 & 87.73 & 98.29 & 94.16 & 97.03 & 87.73 & 96.41 & 75.89 & 97.00 & 77.19\\
		\textbf{SwinFace} & 96.09 & 99.67 & 97.21 & 98.27 & 91.41 & 87.24 & 98.96 & 93.78 & 96.91 & 87.30 & 96.14 & 74.72 & 96.85 & 77.08\\
		\midrule
		& \rotatebox{90}{Receding Hairline} & \rotatebox{90}{Rosy Cheeks} & \rotatebox{90}{Sideburns}
		& \rotatebox{90}{Smiling} & \rotatebox{90}{Straight Hair} & \rotatebox{90}{Wavy Hair} & \rotatebox{90}{Wearing Earrings} & \rotatebox{90}{Wearing Hat} & \rotatebox{90}{Wearing Lipstick} & \rotatebox{90}{Wearing Necklace} & \rotatebox{90}{Wearing Necktie} & \rotatebox{90}{Young} & \rotatebox{90}{} & \rotatebox{90}{\textbf{Average}}\\
		\midrule
		PANDA-1\cite{PANDA-1} & 85.00 & 87.00 & 93.00 & 92.00 & 69.00 & 77.00 & 78.00 & 96.00 & 93.00 & 67.00 & 91.00 & 84.00 & & 85.43\\
		LNets+ANet\cite{CelebA} & 89.00 & 90.00 & 96.00 & 92.00 & 73.00 & 80.00 & 82.00 & 99.00 & 93.00 & 71.00 & 93.00 & 87.00 & & 87.33\\
		MOON\cite{MOON} & 93.56 & 94.82 & 97.59 & 92.60 & 82.26 & 82.47 & 89.60 & 98.95 & 93.93 & 87.04 & 96.63 & 88.08 & & 90.94\\
		NSA\cite{NSA} & 92.25 & 94.79 & 97.17 & 92.70 & 80.41 & 81.70 & 89.44 & 98.74 & 93.21 & 85.61 & 96.05 & 88.01 & & 90.61\\
		MCNN-AUX\cite{MCNN-AUX} & 93.81 & 95.16 & 97.85 & 92.73 & 83.58 & 83.91 & 90.43 & 99.05 & 94.11 & 86.63 & 96.51 & 88.48 & & 91.29\\
		MCFA\cite{MCFA} & 94.00 & 95.00 & 98.00 & 93.00 & 85.00 & 85.00 & 90.00 & 99.00 & 94.00 & 88.00 & 97.00 & 88.00 & & 91.23\\
		DMM-CNN\cite{DMM-CNN} & 94.12 & 95.32 & 97.91 & 93.22 & 84.72 & 86.01 & 90.78 & 99.12 & 94.49 & 88.03 & 97.15 & 88.98 & & \textbf{91.70}\\
		\textbf{SwinFace} & 93.92 & 94.96 & 97.75 & 93.18 & 84.73 & 85.57 & 89.87 & 99.19 & 94.07 & 86.72 & 96.97 & 89.05 & & \underline{91.32}\\
		\bottomrule
	\end{tabular}
	\label{tab:attribute}
\end{table*}

\subsubsection{Face Attribute Estimation}
\begin{figure}[t]
	\centering
	\includegraphics[width=50mm]{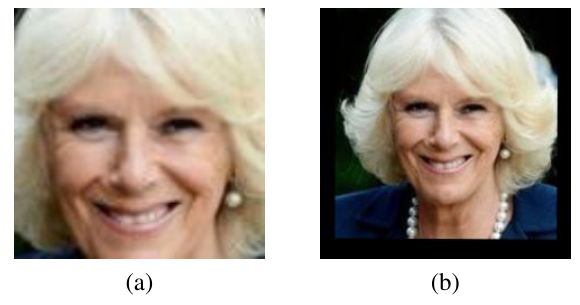}
	
	\caption{(a) Face recognition alignment, which will lose part of the hair and neck information, and is used for datasets other than CelebA\cite{CelebA}. (b) Alignment adopted for CelebA\cite{CelebA}.}
	\label{fig:fig3}
\end{figure}

As shown in \cref{tab:attribute}, We evaluate the face attribute estimation performance on CelebA\cite{CelebA}.
CelebA contains some attributes of hair and neck.
Since we do not want to lose these parts of information, the images cannot be aligned in the way of face recognition as shown in \cref{fig:fig3}(a).
The model can still achieve an average accuracy of 91.32\% comparable to other state-of-the-art methods, proving that the model has excellent generalization ability for faces of different scales.

\subsection{Ablation Study}
\label{ablation}

\begin{table}\footnotesize
	\caption{Comparison for multi-task and single-task learning.}
	\centering
	\begin{tabular}{cccc}
		\toprule
		\multicolumn{1}{c}{\multirow{3}{*}{Setting}}& Age & Smiling & Young\\
		& $\epsilon$-error on  & Acc. on  & Acc. on  \\
		& CLAP2015\cite{CLAP2015} val & CelebA\cite{CelebA} & CelebA\cite{CelebA}\\
		\midrule
		Single-task & 0.357 & 92.40 & 88.38 \\
		Multi-task & \textbf{0.318} & \textbf{93.18} & \textbf{89.05}\\
		\bottomrule
	\end{tabular}
	\label{tab:Ablation1}
\end{table}

\begin{table}\footnotesize
	\caption{ Comparison for different initialization.  }
	\centering
	\begin{tabular}{ccc}
		\toprule
		\multicolumn{1}{c}{\multirow{2}{*}{Initialization}} & Expression Acc. on& Age $\epsilon$-error on\\
		& RAF-DB\cite{DLP-CNN-RAF-DB} & CLAP2015\cite{CLAP2015} val\\
		\midrule
		General recognition  & 86.54 & 0.382\\
		Face recognition & \textbf{91.13} & \textbf{0.357}\\
		\bottomrule
	\end{tabular}
	\label{tab:Ablation2}
\end{table}

\begin{table}\footnotesize
	\caption{ Comparison for Multi-Level Channel Attention module.}
\centering
\begin{tabular}{cccccc}
	\toprule
	\multicolumn{1}{c}{\multirow{3}{*}{Setting}}& Expression & Age & Attribute\\
	& Acc. on & $\epsilon$-error on  & Mean Acc. on  \\
	& RAF-DB\cite{DLP-CNN-RAF-DB} & CLAP2015\cite{CLAP2015} val &CelebA\cite{CelebA} \\
	\midrule
	Baseline & 89.50 & 0.332 & 91.20 &  \\
	MLFF  & 90.51 & 0.336 & \textbf{91.38} \\
	MLFF + CA & \textbf{90.97} & \textbf{0.318} & 91.32  \\
	\bottomrule
\end{tabular}
\label{tab:Ablation3}
\end{table}

\begin{figure}[t]
\centering
\includegraphics[width=80mm]{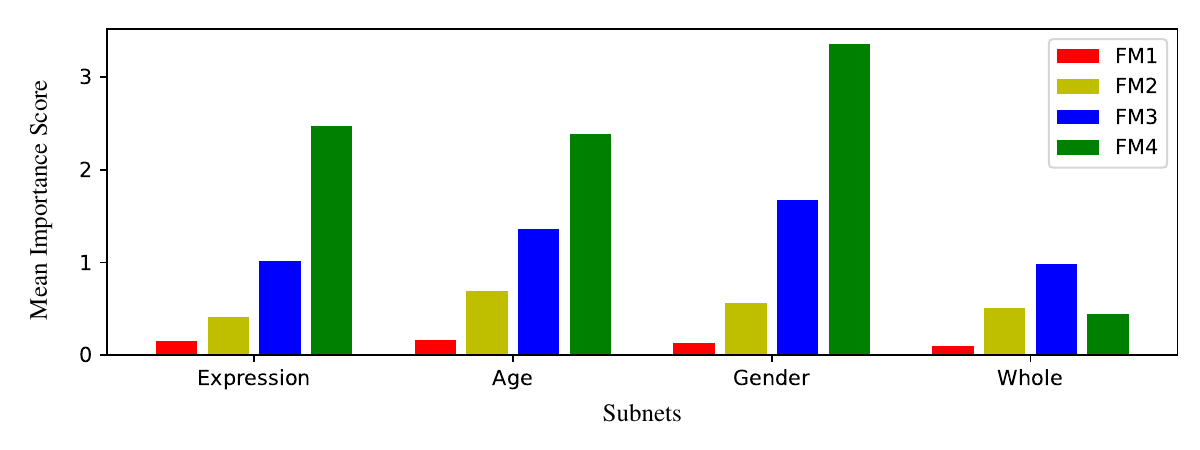}

\caption{Importance of feature maps from different levels for expression, age, gender, and whole face attribute subnets.}
\label{fig:fig4}
\end{figure}

\subsubsection{Multi-task Framework}
\cref{tab:Ablation1} compares the performance of single-task and multi-task training for analysis tasks.
The Swin Transformer backbone is first pre-trained on MS-Celeb-1M\cite{MS-Celeb-1M}.
For simplicity, when evaluating on the CLAP2015 validation set, we do not perform fine-tuning on the CLAP2015 training set.
The experimental results demonstrate the effectiveness of the multi-task learning framework.
Compared with single-task learning, multi-task learning significantly provides superior results of age estimation with a $\epsilon$-error margin of 0.039.
Sharing parameters of the subnet with facial expression recognition (age estimation) task, the attribute classification task for ``Smiling" (``Young") also achieves an increased accuracy by 0.78 (0.67).
We believe that the multi-task learning framework can effectively explore inter-task synergy and learn the correlation among data from different distributions.

\subsubsection{Model Initialization from Face Recognition Task}
ImageNet-1K and MS-Celeb-1M are among the most popular datasets for general recognition and face recognition respectively.
We report the final performances of models pre-trained on ImageNet-1K and MS-Celeb-1M for facial expression recognition and age estimation in \cref{tab:Ablation2}.
The pre-trained model is finetuned on RAF-DB\cite{DLP-CNN-RAF-DB} and AffectNet\cite{AffectNet} (IMDB+WIKI\cite{IMDB+WIKI}, Adience\cite{Adience} and MORPH\cite{MORPH}) for facial expression recognition (age estimation).
For simplicity, when evaluating on the CLAP2015 validation set, we do not further perform fine-tuning on the CLAP2015 training set.
Results show that the model initialization from face recognition can significantly improve the performance of analysis tasks lacking large-scale clean labels.
The $\epsilon$-error on the CLAP2015\cite{CLAP2015} valuation set decreases by 0.025 and the accuracy on RAF-DB\cite{DLP-CNN-RAF-DB} increases by 4.6.

\subsubsection{Multi-Level Channel Attention}
\cref{tab:Ablation3} shows the results of three networks.
In the baseline network, analysis subnets only use the feature map from the top layer of the backbone.
It can be found that using both MLFF and CA can improve the performance of age estimation, face attribute estimation, and facial expression recognition.
This shows that MLCA can effectively alleviate the feature extraction conflict of the backbone while adaptively selecting robust feature representations for subnets.
Among them, FER benefits most from the MLCA mechanism, with a particularly obvious accuracy increase of 1.47\% on RAF-DB\cite{DLP-CNN-RAF-DB}.
This makes sense, as the target conflict between FER and FR is the most serious.

\subsubsection{Importance of Feature Maps for Different Subnets}
We want to know which level of the feature maps the subnets prefer.
In a face analysis subnet, the channel-weighted feature map is passed through a max pooling layer and a ReLU activation layer to obtain a 512-dimensional vector.
The activation values in the 512-dimensional vector can represent the importance scores of the corresponding channels in the inference phase.
As shown in \cref{fig:fig4}, we average the importance scores of channels from FM1, FM2, FM3, and FM4.
In general, a feature map from deeper layers gets a higher importance score.
It is worth noting that the feature map from the top layer is not always the most useful.
The whole face attribute subnet handles six estimation tasks of ``Attractive", ``Blurry", “Chubby”, ``Heavy Makeup", ``Oval Face", and ``Pale Skin", as shown in \cref{tab:subnets}.
For this subnet, the third feature map is more beneficial.

\subsubsection{Loss Function Selection and Subnet Design for Face Recognition}
To determine the loss function and subnet design details for face recognition, we conduct experiments on CASIA-WebFace\cite{CASIA-WebFace} which contains 494K images of 10,572 celebrities.
\cref{tab:selection} reports the performance with different loss functions and subnet designs.
For Swin Transformer-based face recognition, CosFace\cite{CosFace} loss function can provide better performance than ArcFace, which is commonly used for CNN-based models.
The introduced FC-BN-FC-BN structure also contributes to performance improvement.

\begin{table}\small
\caption{Comparison for face recognition using different loss functions and subnet designs.}
\centering
\begin{tabular}{c|c|ccc}
	\toprule
	Loss & FC-BN-FC-BN & LFW & CFP-FP & AgeDB-30 \\
	\midrule
	ArcFace & without & 95.68 & 88.44 & 78.85 \\
	CosFace & without & 98.80 & 90.66 & 90.13 \\
	CosFace & with & \textbf{98.97} & \textbf{91.66} & \textbf{90.23} \\
	\bottomrule
\end{tabular}
\label{tab:selection}
\end{table}

\subsubsection{Model Running Efficiency}
We evaluate the running efficiency of the proposed SwinFace on one NVIDIA Tesla T4 GPU and the batch size is set to 32.
It takes an average of 3.205ms to calculate the recognition feature using the Swin Transformer backbone and face recognition subnet, while it takes an average of 4.357ms to gain all 43 outputs, which increases the time overhead by 36\%.
The result shows that our method can significantly improve the overall application efficiency of face recognition and analysis.

\section{Conclusion}

This paper proposes a multi-task face recognition and analysis model based on Swin Transformer\cite{Swin}, which can perform face recognition, facial expression recognition, age estimation, and face attribute estimation simultaneously.
Extensive experiments show that our method can lead to a better understanding of faces.
The proposed MLCA module allows analysis subnets to acquire features from different levels of the backbone and adaptively find the features from appropriate levels, improving performance.
In addition, our work emphasizes the importance of robust initialization from face recognition for facial expression recognition and age estimation and achieves SOTA on both tasks.
The current model still has limitations in terms of its functionality.
In the future, we plan to extend our model for face localization tasks, such as head pose estimation, face alignment, face parsing, and 3D face reconstruction. 
Besides, although not validated in our work, an iterative pseudo-labeling process for semi-supervised learning could potentially further enhance model performance in tasks with limited labeled data, such as age estimation and facial expression recognition.

\section*{Acknowledgments}
This work was funded by the Beijing University of Posts and Telecommunications-China Mobile Research Institute Joint Innovation Center, National Natural Science Foundation of China under Grant No. 62236003, and China Postdoctoral Science Foundation under Grant 2022M720517. 
This work was also supported by Program for Youth Innovative Research Team of BUPT No. 2023QNTD02.



\vspace{11pt}

\begin{IEEEbiography}[{\includegraphics[width=1in,height=1.25in,clip,keepaspectratio]{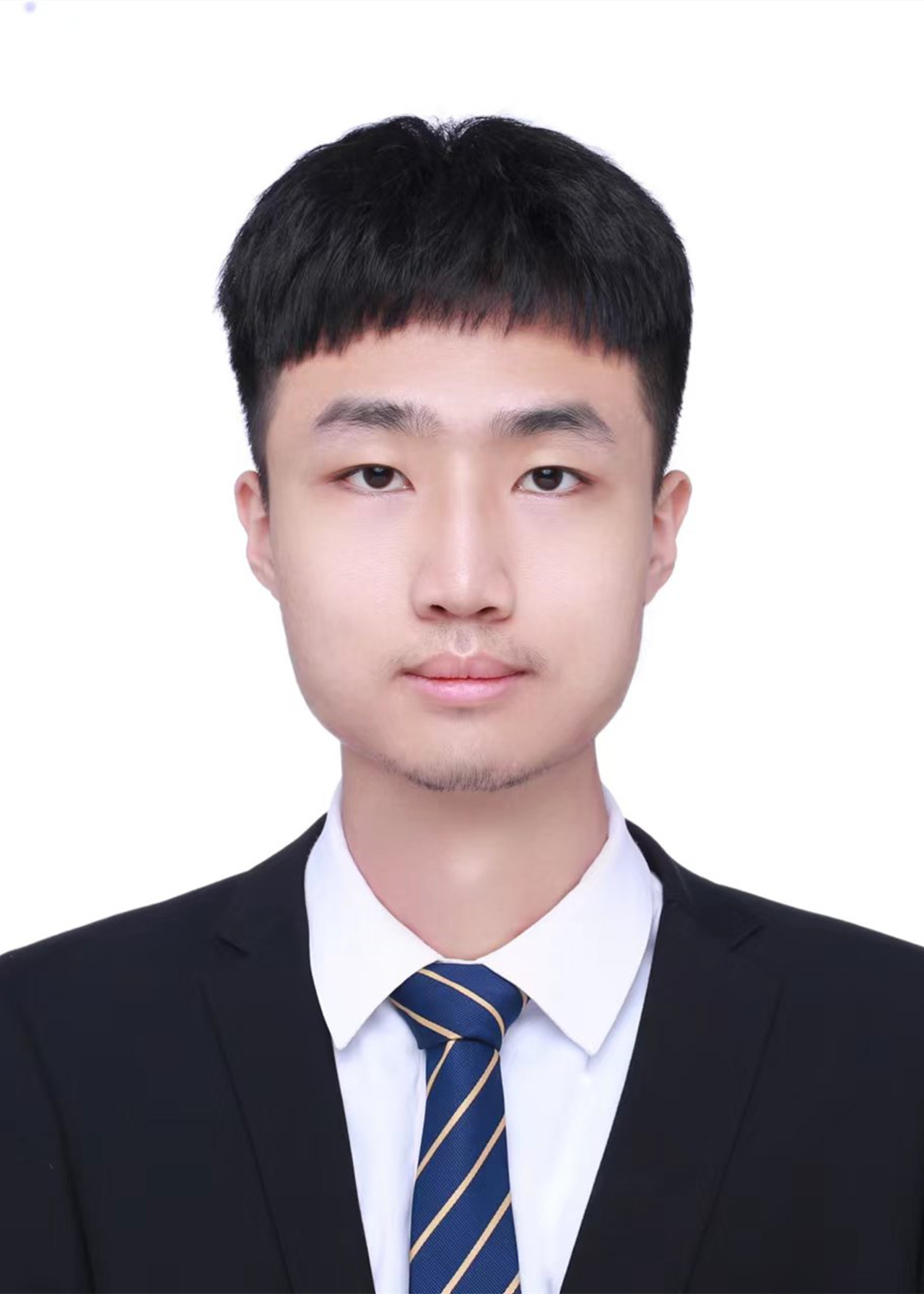}}]{Lixiong Qin} received the B.E. degree from Beijing University of Posts and Telecommunications (BUPT), Beijing, China, in 2022, where he is currently pursuing the master’s degree in artificial intelligence with the School of Artificial Intelligence. 
His research interests include computer vision and face analysis.
\end{IEEEbiography}

\vspace{11pt}
\begin{IEEEbiography}[{\includegraphics[width=1in,height=1.25in,clip,keepaspectratio]{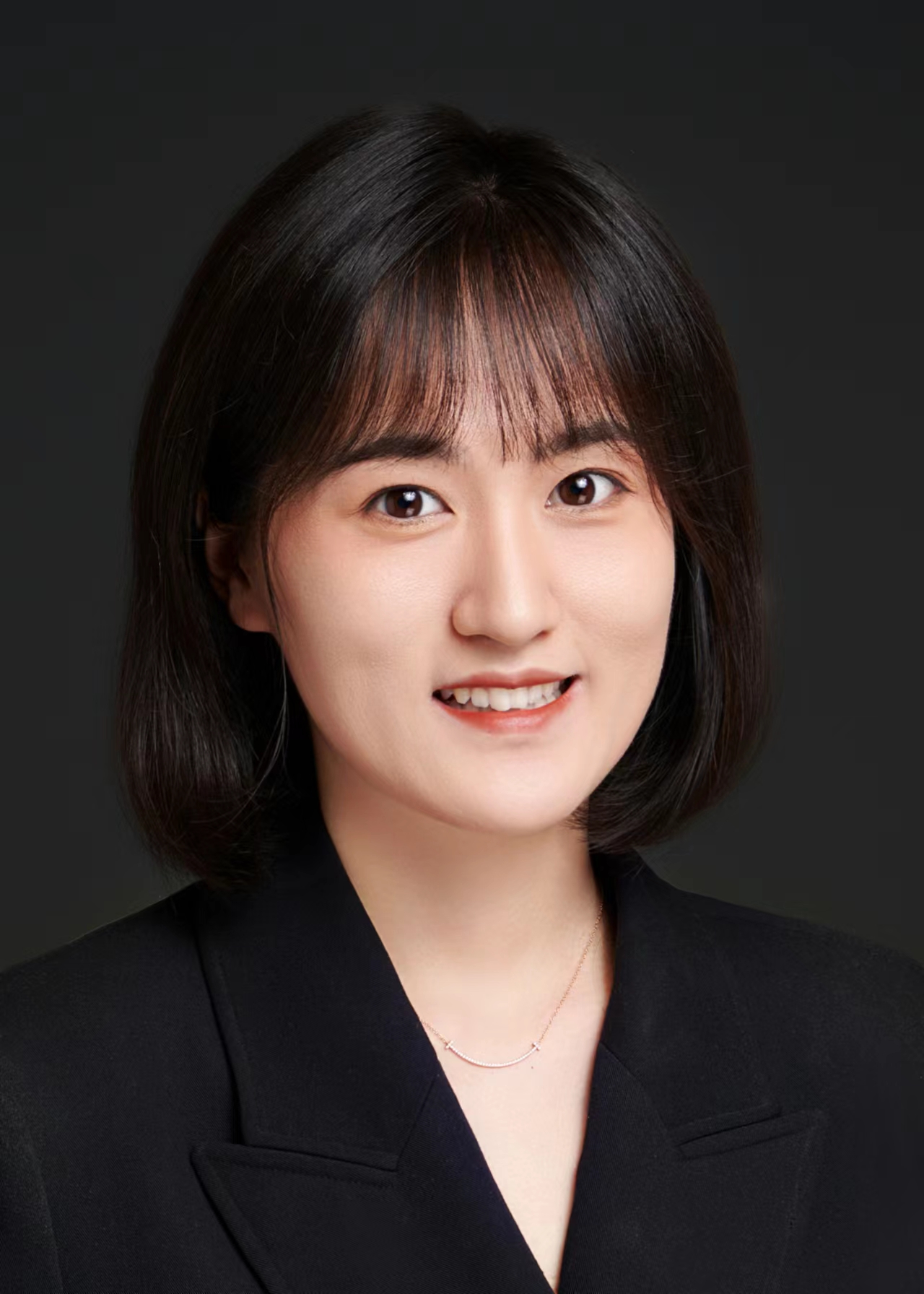}}]{Mei Wang} received the B.E. degree and Ph.D. degree in information and communication engineering from Beijing University of Posts and Telecommunications (BUPT), Beijing, China, in 2013 and 2022, respectively. She is currently a Postdoc in the School of Artificial Intelligence, Beijing University of Posts and Telecommunications, China. Her research interests include computer vision, with a particular emphasis in computer vision, domain adaptation and AI fairness.
\end{IEEEbiography}

\vspace{11pt}
\begin{IEEEbiography}[{\includegraphics[width=1in,height=1.25in,clip,keepaspectratio]{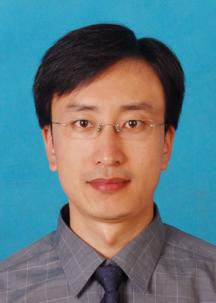}}]{Chao Deng} received the M.S. degree and the Ph.D. degree from Harbin Institute of Technology, Harbin, China, in 2003 and 2009 respectively. He is currently a deputy general manager with AI center of China Mobile Research Institute. His research interests include machine learning and artificial intelligence for ICT operations.
\end{IEEEbiography}

\vspace{11pt}
\begin{IEEEbiography}[{\includegraphics[width=1in,height=1.25in,clip,keepaspectratio]{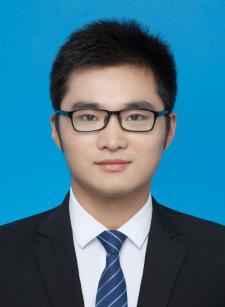}}]{Ke Wang} received the M.S. degree form Beijing Institute of Technology in 2018. He is currently an algorithm engineer in China Mobile Research Institute. His research interests include 3D reconstruction, 3D face recognition and multi-modal fusion.
\end{IEEEbiography}

\vspace{11pt}
\begin{IEEEbiography}[{\includegraphics[width=1in,height=1.25in,clip,keepaspectratio]{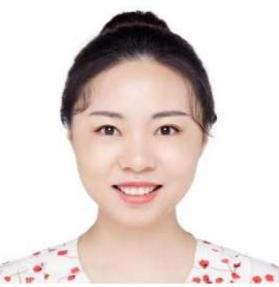}}]{Xi Chen} received her M.S. degree from Communication University of China. She works as an algorithm engineer in image and video processing and computer vision at China Mobile Research Institute since 2018.
\end{IEEEbiography}

\vspace{11pt}
\begin{IEEEbiography}[{\includegraphics[width=1in,height=1.25in,clip,keepaspectratio]{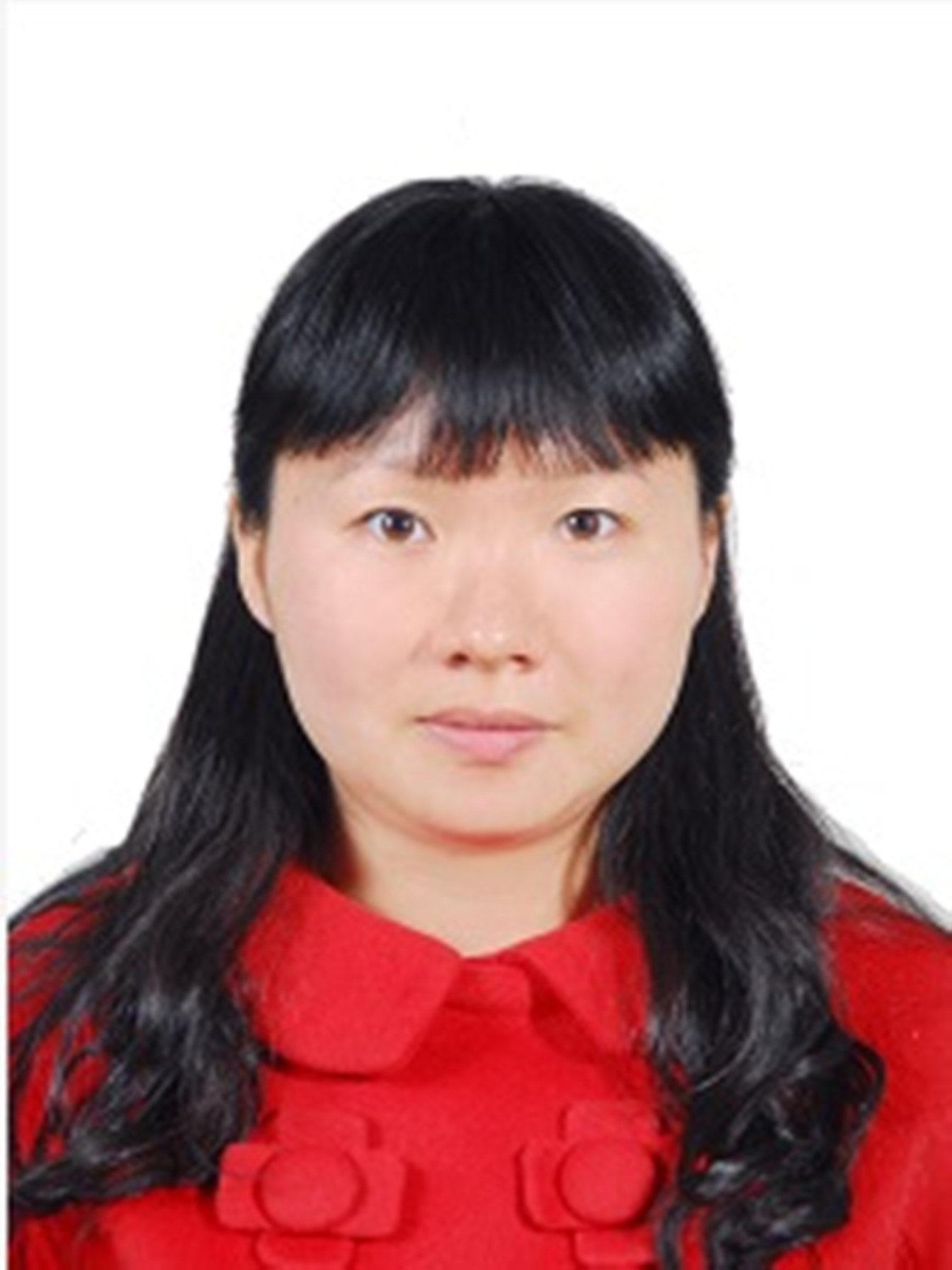}}]{Jiani Hu} received the B.E. degree in telecommunication engineering from China University of Geosciences in 2003, and the Ph.D. degree in signal and information processing from Beijing University of Posts and Telecommunications (BUPT), Beijing, China, in 2008. She is currently an associate professor in School of Artificial Intelligence, BUPT. Her research interests include information retrieval, statistical pattern recognition and computer vision.
\end{IEEEbiography}

\vspace{11pt}
\begin{IEEEbiography}[{\includegraphics[width=1in,height=1.25in,clip,keepaspectratio]{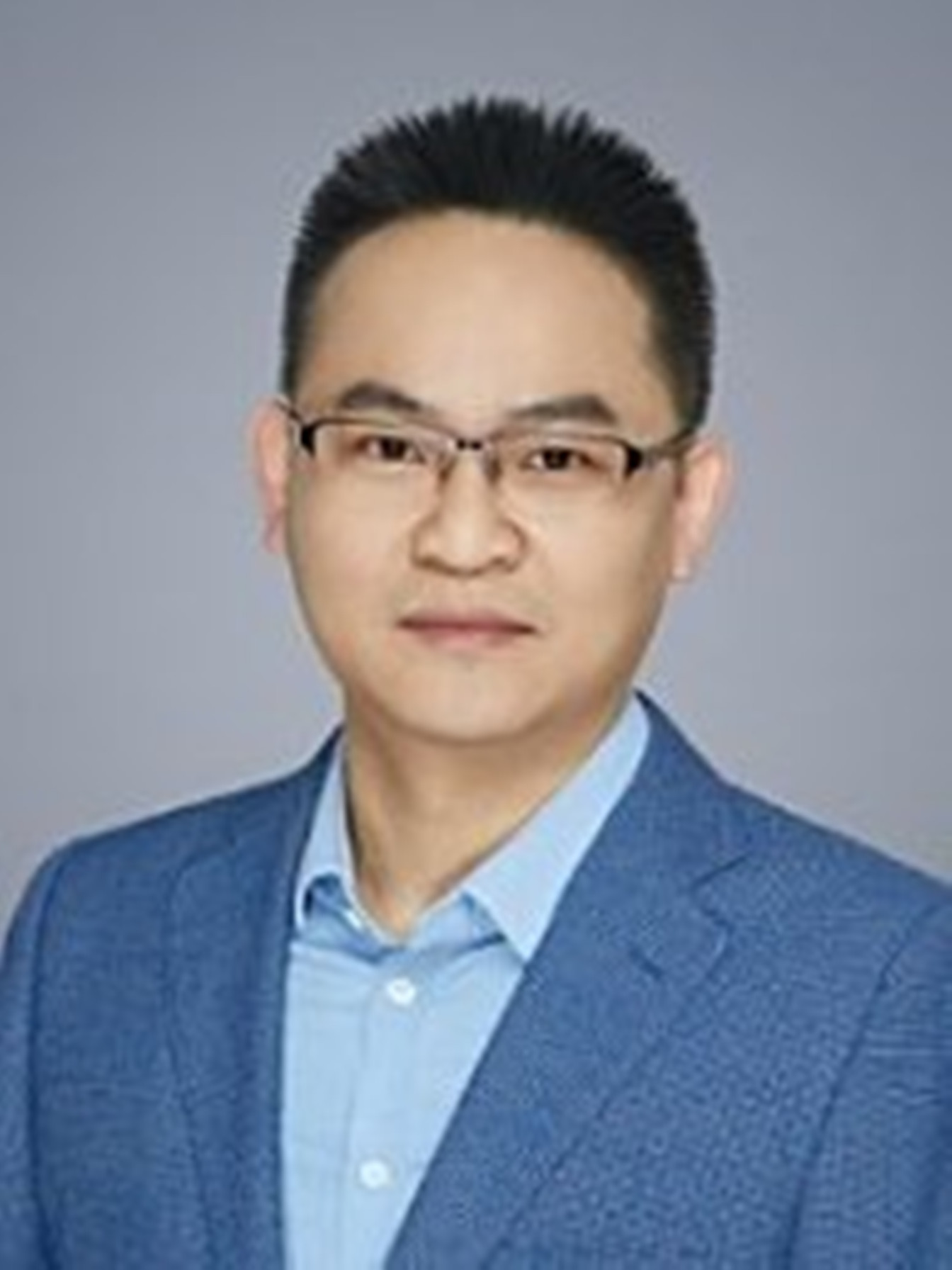}}]{Weihong Deng} received the B.E. degree in information engineering and the Ph.D. degree in signal and information processing from the Beijing University of Posts and Telecommunications (BUPT), Beijing, China, in 2004 and 2009, respectively. He is currently a professor in School of Artificial Intelligence, BUPT. His research interests include trustworthy biometrics and affective computing, with a particular emphasis in face recognition and expression analysis. He has published over 100 papers in international journals and conferences, such as IEEE TPAMI, TIP, IJCV, CVPR and ICCV. He serves as area chair for major international conferences such as IJCB, FG, IJCAI, ACMMM, and ICME, guest editor for IEEE Transactions on Biometrics, Behavior, and Identity Science, and Image and Vision Computing Journal, and the reviewer for dozens of international journals and conferences. His Dissertation was awarded the outstanding doctoral dissertation award by Beijing Municipal Commission of Education. He has been supported by the programs for New Century Excellent Talents and Young Changjiang Scholar by Ministry of Education.
\end{IEEEbiography}

\vfill

\end{document}